\documentclass[journal]{IEEEtran}
\let\emph\textit
\usepackage{cite}
\usepackage{threeparttable}
\usepackage{amssymb,amsfonts}
\usepackage{algorithmic}
\usepackage{amsmath} 
\usepackage{autobreak}
\usepackage{graphicx}
\usepackage{textcomp}
\usepackage[numbers]{natbib}
\usepackage{algorithm}
\usepackage{enumerate}
\usepackage{subfigure}
\usepackage{multirow}
\usepackage{autobreak}
\usepackage{booktabs}
\usepackage{color, soul}
\usepackage{url}
\usepackage{dutchcal}

\usepackage[T1]{fontenc}

\ifCLASSINFOpdf
\else
\fi
\usepackage{amsmath}
\usepackage[cmintegrals]{newtxmath}
\begin{document}

%
\title{Multi-Objective Evolutionary for Object Detection Mobile Architectures Search}

%
%
%
\author{Haichao Zhang\#,
Jiashi Li,
Xin Xia,
Kuangrong Hao*,
Xuefeng Xiao
\thanks{This work was supported in part by the Fundamental Research Funds for the Central Universities (2232021A-10, 2232021D-37), National Natural Science Foundation of China (61903078), Natural Science Foundation of Shanghai (20ZR1400400, 21ZR1401700), and the Fundamental Research Funds for the Central Universities and Graduate Student Innovation Fund of Donghua University (CUSF-DH-D-2021048).}
\thanks{Corresponding author: Kuangrong Hao}
\thanks{H.-C. Zhang, K. Hao are now with the College of Information Science and Technology, Engineering Research Center of Digitized Textile and Apparel Technology, Ministry of Education, Donghua University, Shanghai 201620, P. R. China (e-mail: krhao@dhu.edu.cn).}
\thanks{J.-S. Li, X. X., and X.-F. X are now with ByteDance, Intelligent Creation(e-mail:\{lijiashi, xiaxin.97,xiaoxuefeng.ailab\}@bytedance.com)}
\thanks{\#This work was done when the first author was an intern at ByteDance.}
}

\maketitle

\begin{abstract}
Recently, Neural architecture search has achieved great success on classification tasks for mobile devices. The backbone network for object detection is usually obtained on the image classification task. However, the architecture which is searched through the classification task is sub-optimal because of the gap between the task of image and object detection. As while work focuses on backbone network architecture search for mobile device object detection is limited, mainly because the backbone always requires expensive ImageNet pre-training. Accordingly, it is necessary to study the approach of network architecture search for mobile device object detection without expensive pre-training. In this work, we propose a mobile object detection backbone network architecture search algorithm which is a kind of evolutionary optimized method based on non-dominated sorting for NAS scenarios. It can quickly search to obtain the backbone network architecture within certain constraints. It better solves the problem of suboptimal linear combination accuracy and computational cost. The proposed approach can search the backbone networks with different depths, widths, or expansion sizes via a technique of weight mapping, making it possible to use NAS for mobile devices detection tasks a lot more efficiently. In our experiments, we verify the effectiveness of the proposed approach on YoloX-Lite, a lightweight version of the target detection framework. Under similar computational complexity, the accuracy of the backbone network architecture we search for is $2.0\%$ mAP higher than MobileDet. Our improved backbone network can reduce the computational effort while improving the accuracy of the object detection network. To prove its effectiveness, a series of ablation studies have been carried out and the working mechanism has been analyzed in detail.
\end{abstract}


\section{Introduction}

Object detection is widely used in many computer vision tasks, including image annotation, object tracking, segmentation, and human activity recognition, with a wide range of applications, such as autonomous driving, drone obstacle avoidance, robot vision, human-computer interaction, and augmented reality \cite{xiao2017design,xiao2017building,li2022sepvit}. Deep neural network-based object detection has received a lot of attention from academia and industry. In many computer vision applications, it can be observed that higher capacity networks lead to higher performance \cite{ren2015faster,fang2021you,zhang2019freeanchor}. However, they are often more resource-consuming. This makes it challenging to find models with the right quality-compute trade-off for deployment on edge devices with limited inference budgets. Therefore, designing object detection neural network architectures for efficient deployment on mobile devices is not an easy task: the computational volume and accuracy must be wisely traded off.

Image classification has always been a fundamental task in the design of neural structures. Typically, networks designed and pre-trained on classification tasks are utilized as the backbone and fine-tuned to split or detect tasks. However, the backbone plays an important role in these tasks, and the differences between these tasks require different backbone design principles. For example, target detection tasks require moving positioning and classification prediction from each convolution feature. This difference makes the neural architecture designed for categorical tasks inadequate. Some tentative work \cite{li2018detnet,wang2020deep} has been done to address this issue by manually modifying the architecture designed for classification to better accommodate the characteristics of new tasks. Similarly, some efforts have been devoted to the manual design of lightweight neural network architectures for mobile devices \cite{wang2018pelee,lin2020mcunet,wang2020glance}. Unfortunately, relying on human expertise is time-consuming and potentially suboptimal. 

Correspondingly, many methods have been proposed to solve the optimization requirements of lightweight neural network architecture.  In particular, Neural Architecture Search (NAS) \cite{white2021powerful,Xia_2022_WACV,li2022next,xia2022trt} provides a framework for automating the design of neural network architectures for mobile devices. To address the need for automatic tuning of neural network structures, NAS \cite{howard2017mobilenets} has shown excellent capabilities in searching network architecture that are not only accurate but also efficient on specific hardware platforms. Automatic depth learning is designed to help engineers avoid a lot of experimentation and errors in architecture design and further improve the performance of architecture over artificially designed architectures. Early NAS work \cite{tan2019mnasnet,cai2018proxylessnas} explored search issues on classification tasks. With the development of NAS methods, some work \cite{zhang2019customizable,chen2019detnas} has proposed the use of NAS to specialize the backbone architecture design of object detection tasks.

Likewise, aiming at object detection tasks for mobile devices, a lot of efforts have been made in neural architecture search. In early work, NAS-FPN \cite{ghiasi2019fpn} is proposed, which searches for feature pyramid networks (FPNs) instead of a backbone. It can be executed on a pre-trained backbone network and searched using previous NAS algorithms. As a result, the difficulty of trunk search remains unresolved. MobileDet \cite{RN1282} is proposed to solve the optimization problem of backbone network architecture when edge devices cannot use deep separable convolution. MnasFPN \cite{chen2020mnasfpn} designs a search space with good detection heads for mobile and combines it with delay-aware architecture search to generate efficient object detection models. Based on the basic idea of Darts, FNA \cite{RN1152} automates the design of the backbone network of the object detection network by means of parameter remapping in the MobileNetv2-based search space. However, these methods optimize the computational complexity and accuracy of the network by combining the indicators of both linearly. According to the work \cite{deb2002fast}, the linear combination of objectives is sub-optimal. Therefore, it is necessary to embed this search problem into a realistic multi-objective environment where a series of models are found along the Pareto front, such as accuracy and computational cost.

Our work aims to reconsider the optimization of mobile backbone networks based on Single path one-shot neural architecture search (SPOS) and design an evolutionary-based method through non-dominated sorting for NAS scenarios. To take full advantage of the SPOS and evolutionary-based method through non-dominated sorting, \textbf{Mobile Non-dominated Sorting Genetic Algorithm Neural Architecture Search} (MNSGA-NAS) is proposed. A GhostNet-based search space cluster is constructed and a YoloX-based \cite{ge2021yolox} search framework is utilized for object detection tasks. MNSGA-NAS follows the idea of SPOS, and a supernet is pre-trained on ImageNet and  fine-tuned on COCO with the YoloX \cite{ge2021yolox}. Therefore, the fine-tuned supernet has the performance conditions to evaluate the backbone architecture on COCO. In order to search for backbone networks with different depths, widths, or scales, we design a weighting mapping technique, which implements an individual evaluation based on supernet, so as to search for backbone networks for mobile device detection tasks more effectively. Then, we design the evolutionary approach based on non-dominated sorting for searching the backbone network and obtain the backbone network architecture to be searched. To evaluate the performance of the proposed approach, in the same search space, we compare the backbone network (SPOS CLS.) oriented to the classification task through SPOS and the backbone network through MNSGA-NAS. The models searched by MNSGA-NAS outperform the SPOS CLS.+YoloX Lite by 1.1 mAP on COCO test-dev as comparable FLOPs. The results report that the accuracy of the backbone searched for detection is superior to that of the classification-oriented task. MNSGA-NAS also outperforms MobileDets-IBN \cite{RN1282} by 3.2 mAP as comparable FLOPs. In addition, the models searched by MNSGA-NAS are converted to TensorRT by MMdeploy \cite{mmdeploy} and verified the inference performance on Jetson TX2 NX, and the results obtained competitive performance.

Our main contributions can be summarized as follows:

\begin{itemize}
  \item An evolutionary-based approach with non-dominated ranking is proposed for the lightweight backbone network architecture search. The approach can quickly construct Pareto surfaces by reconstructing the non-dominant ordering to obtain a rapidly deployable backbone on mobile devices for detection.
  \item We design a weight mapping technique to search for backbone networks with different depths, widths, or extensions. This proposed method takes full advantage of SPOS and pruning to improve the accuracy of backbone networks as much as possible with limited FLOPs.
  \item The searched models consume fewer computing resources and are friendly to mobile devices. These models can be quickly deployed to mobile devices such as Jetson and achieve competitive performance.
\end{itemize}
\section{Related Work}
\begin{figure*}
  \centering  
  \includegraphics[width=14cm]{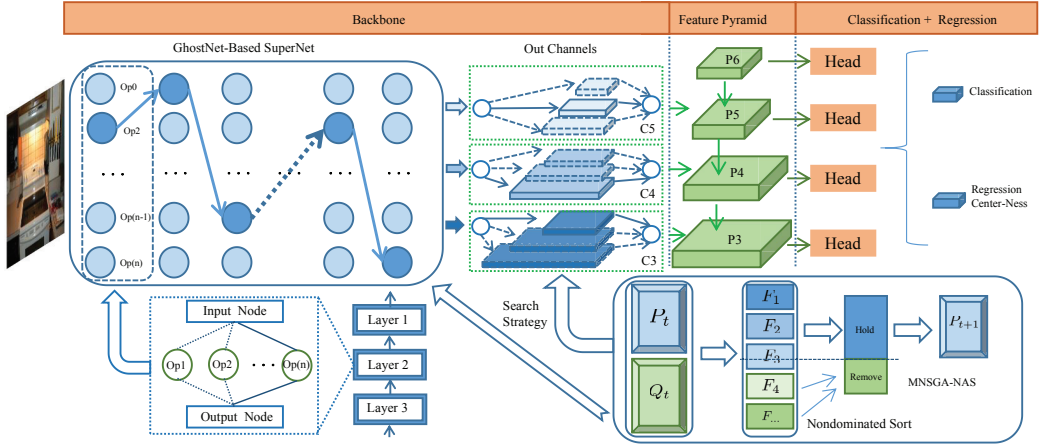}
  \caption{\footnotesize The framework of our proposed MNSGA-NAS. The GhostNet-Based Supernet is pre-trained on ImageNet and fine-tuned on COCO. In the searching process, the supernet is sampled as a single path one-shot approach with uniform sampling. The depth of the network is controlled by the number of identity operations. The number of output channels is searched by the pruning as the L1 norm of the channels weights. We perform channel sorting and pick the important channels(with the large L1 norm) to map the smaller numbers channels. The network depth, network structure, and network output channels are encoded to be searched by the evolution-based  approach based on non-dominated sorting for detection-specific NAS.}
  \label{fig_1}
  \vspace{-2.0em}
\end{figure*}

\subsection{Multi-objective NAS}

Deploying neural network models on a mobile scene requires a multi-objective perspective to complete the deployment and achieve a balance between accuracy and complexity. MONAS \cite{hsu2018monas} optimizes the prediction accuracy and other objectives by extending NAS through planning a linear combination. However, according to \cite{deb2002fast}, the linear combination of objectives is suboptimal. Therefore, it is necessary to embed the architecture search problem into a real multi-objective environment where a series of network architectures are found along the Pareto frontier of multiple objectives, such as accuracy, computational cost or inference time, etc. LEMONADE \cite{elsken2018efficient} utilizes the Lamarckian inheritance mechanism to generate child individuals from parents. NEMO \cite{Kim2017NEMON} and NSGA-Net \cite{RN994} utilize the classical non-dominated search algorithm (NSGAII) to handle trade-offs between accuracy and complexity. The group models are based on dominance while prioritizing models within the same front when measuring the crowding distance. However, the above algorithm only considers the architectural properties of the models for the classification task and fails to extend the task to the optimization of the backbone network for object detection.

\subsection{Mobile Neural Architecture Search (NAS)}

Earlier, mobile neural network architecture ware designed by handcrafted, such as the family of MobileNet \cite{howard2017mobilenets,sandler2018mobilenetv2} and the family of ShuffleNet \cite{ma2018shufflenet,zhang2018shufflenet}. Then, with the widespread interest in AutoML, neural architecture search (NAS) is utilized in the design of the mobile neural network in a proper search space. NetAdapt \cite{yang2018netadapt} and AMC \cite{he2018amc} were the first attempts to fine-tune the number of channels of pre-trained models to accelerate model inference. Moreover, MobileNet-v3 \cite{howard2019searching} search the resource-efficient architecture within the NAS framework and obtained the state-of-the-art mobile neural network architectures. 

Currently, most of the literature \cite{tan2019mnasnet,tan2019efficientnet} of NAS consistently focuses on improving the search efficiency on classification tasks and only utilize the learned feature extractor as a backbone for target detection, which has not been further explored in object detection. In addition, \cite{RN1152,wang2020fcos,RN1282,chen2020mnasfpn} have shown that a better complexity-accuracy tradeoff can be obtained by directly searching the architecture of the target detection network. MnasFPN \cite{chen2020mnasfpn} is a detection-faced NAS framework for mobile object detection networks, which searches the feature pyramid head in a mobile-friendly search space. Yet, some factors limit its mobile generalize, such as MnasFPN does not search for the backbone which is one of the object detection performance bottlenecks. MobileDets \cite{RN1282} is proposed to search for a better detection backbone network for mobile with high real-time requirements. It proposes a new search space that takes into account the impact of different platforms (CPUs, EdgeTPUs, DSPs) and yields a backbone suitable for detection. However, it has shortcomings in the mixed search of the depth and the number of output channels of the BackBone. By comparison, our work relies on YoloX \cite{ge2021yolox} head which is more amenable to mobile acceleration. We perform a new encoding of the search space and obtain an efficient search of the target detection backbone network based on the proposed novel multi-objective evolutionary algorithm.

\section{Methods}

In this section, we first explain the pipeline of the proposed approach, which is performed on ImageNet pre-trained and COCO fine-tuned. Then, we describe the evolving approach to architecture search scenarios. At the same time, we describe the weight mapping approach we have designed for individual performance evaluation during evolution. Finally, we describe the search space for the proposed approach.

\subsection{Search Pipeline}

As the idea of MNSGA-NAS follows the approach of SPOS \cite{guo2020single}, the MNSGA-NAS pipeline consists of 4 steps: backbone supernet is pre-trained on ImageNet, backbone supernet with detection framework is fine-tuned on COCO, the architecture of backbone is searched on the supernet with detection framework, and the whole detection network is trained without ImageNet pre-trained.

\textbf{Step 1: Backbone supernet pretraining}. ImageNet pre-trained is the fundamental step of the supernet network construct. From the one-shot approach \cite{guo2020single}, the discrete GhostNet-based search space is relaxed into a continuous one, which makes the weights of individual networks deeply coupled. In the process of training supernet, only one single path is uniformly sampled for feedforward and backward propagation. The gradient update is only utilized for the nodes being sampled.

\textbf{Step 2: Supernet fine-truning} Supernet with ImageNet pre-trained is also sampled in only one single path, but equipped with a detection head, feature fusion pyramid, and detection datasets. Fig~\ref{fig_1} shows the supernet with detection head and feature fusion pyramid. Especially, the BN is infeasible because the task of object detectors is trained with high-resolution images which, unlike image classification, so the batch size constrained small. The small batch size can lead to severely degrading the accuracy of BN. To this end, the convolutional BN is replaced with synchronized Batch Normalization(SyncBN) during supernet fine-tuning. 

\textbf{Step 3: Search on supernet with non-dominated sorting} The third step is to conduct the architecture search on the detection network with fine-turned Supernet. Architecture paths are picked and evaluated under the approach of the non-dominated sorting evolutionary controller. For the evolutionary search process, the details can be shown in Section III-B. During the search, each individual network architecture is sampled as a single path and a fixed channel width in the supernet via the weighting mapping technique. For the weighting mapping technique, the details can be seen in Section III-C. The channel width means the number of network channels. The Supernet is trained with full-width and we prune channels to flexibly implement the network output of the number of channels. In the channels search process, it calculates the channel importance based on the L1 norm of the channel weights, where a larger L1 norm means more importance. Therefore, the channel with a larger L1 norm is left during the search \cite{cai2019once,he2018amc}. On the other hand, the accuracy of the whole network is rapidly reduced due to the pruning of some key channels during pruning channels. Therefore, the whole network is sampled by the path and the number of channels, and appropriate training is performed to achieve rapid recovery of accuracy \cite{han2015deep,lin2017runtime}. The training details for accuracy recovery are shown in Section IV.A-B.

\textbf{Step 4: Training from scrath without ImageNet pretrained} Following the method of training networks in YoloX \cite{ge2021yolox}, the MixUp and Mosaic Implementation are adopted to train the YoloX-based network. In practice, we find that ImageNet pre-training is no more beneficial and often results in overfitting of the network. Therefore, we train all the detection networks from scratch.

\subsection{Search Strategy}

In some multi-objective architecture search methods \cite{RN1282,chen2020mnasfpn,RN1152}, the index linear combination of the two is usually used to optimize the computational complexity and accuracy of the network. However, the linear combination of the object is suboptimal. To overcome the suboptimal problem caused by a linear combination of multi-objective by the loss function during neural architecture search for object detection, we propose a multi-object architecture search algorithm for mobile object detection. Thus, we introduce the proposed multi-objective neural network architecture search strategy based on non-dominated sorting, which is based on NSGA-II and improved for the characteristics of architecture search.

\begin{algorithm}[!htbp]
  \footnotesize
  \caption{\footnotesize Mobile Non-dominated Sorting Genetic Algorithm Neural Architecture Search}
  \label{20220509_1}
  \begin{algorithmic}[1]
    \REQUIRE The population size $ N $ , the maximal generation number $ T $ , the crossover probability $ \mu  $ , and the mutation probability $ \nu  $. Constraint function $g_1$and $g_2$.
    \STATE Randomly initialize the population $ P $  with the size of $ N $ by uniform sampling.
    \STATE $\left[ F_1,F_2,... \right] \gets $ Fast non-dominated sorting $ \left( \mathcal{L}_{val}\left( P \right) ,\mathcal{C}_1\left( P \right) ,\mathcal{C}_2\left( P \right)\right) $.
    \STATE $ \left[ G_1,G_2,... \right] \gets  $ Crowding distance sorting $ \left[ F_1,F_2,... \right]  $.    
    \FOR{ $ t=1,2,3,...,T $ }
      \STATE Clear offspring population $ Q_1 \gets \varnothing  $  ,$ i\gets 0 $.      
        \WHILE{$i< 4N $ or $||Q_1||< N$} 
          \STATE  $p\gets$ Binary Tournament Selection $ \left( P,\left[  G_1,G_2,... \right] ,  \left[ F_1,F_2,... \right] \right)  $ select individual by the tournament.
          \STATE  \textbf{Crossover:} $ q\gets Crossover\left( p,\mu \right)  $.
          \STATE \textbf{Mutatuin:} $ q\gets Mutatuin\left( q,\nu \right) $. 
          \STATE From $q$ get the simple objective $\mathcal{C}_1\left( q \right) $ and $\mathcal{C}_2( q )$ 
          \STATE Choose $q$ based on constraint function $g_1$and $g_2$ through $\mathcal{C}_1\left( q \right) $ and $\mathcal{C}_2( q )$.
          \STATE $ Q_1\gets Q_1 \cup q;i=i+1 $.
        \ENDWHILE             
      \STATE $\left[ F_1,F_2,... \right] \gets $ Fast non-dominated sorting $ \left( \mathcal{C}_1\left( (P\cup Q_1) \right) \right),\mathcal{C}_2\left( (P\cup Q_1) \right) $.
      \STATE $ \left[ G_1,G_2,... \right] \gets  $ Crowding distance sorting $ \left[ F_1,F_2,... \right] $.
      \STATE $P\gets $ Selecttion$((P\cup Q_1),\left[ F_1,F_2,... \right],\left[ G_1,G_2,... \right])$
      \STATE Clear offspring population $ Q \gets \varnothing  $  ,$ i\gets 0 $. 
      \WHILE{$i< N $ } 
          \STATE  $p\gets$ Binary Tournament Selection $ \left( P,\left[ G_1,G_2,... \right] , \left[ F_1,F_2,... \right] \right)  $ select individual by the tournament.
          \STATE  \textbf{Crossover:} $ q\gets Crossover\left( p,\mu \right)  $.
          \STATE \textbf{Mutatuin:} $ q\gets Mutatuin\left( q,\nu \right) $.           
          \STATE $ Q\gets Q \cup q;i=i+1 $.
        \ENDWHILE      
      \STATE $ \left[ F_1,F_2,... \right] \gets  $ Fast non-dominated sorting $ \left( \mathcal{L}_{val}\left( P\cup Q \right) ,\mathcal{C}_1\left( P\cup Q \right) \right),\mathcal{C}_2\left(P\cup Q \right)) $.
      \STATE $ \left[ G_1,G_2,... \right] \gets  $ Crowding distance sorting $ \left[ F_1,F_2,... \right]  $  
      \STATE $ P\gets \left( P\cup Q,\left[ G_1,G_2,... \right] ,\left[ F_1,F_2,... \right] \right)  $ select individual by the tournament.
    \ENDFOR
    \STATE The Pareto population $P$.
\end{algorithmic}
\end{algorithm}

For practical deployment scenarios in mobile devices, we implement network architecture complexity and performance trade-offs from a multi-objective optimization perspective. In this work, we treat the automated design of object detection backbone architecture for the mobile device as a bilevel optimization problem. We mathematically formulate the optimization problem as:

\begin{equation}
  \begin{array}{c}
    \text{minimize} \ \mathbf{F}(\alpha) = (\mathcal{L}_{val}(\omega^*(\alpha),\mathcal{C}_0(\alpha),\mathcal{C}_1(\alpha)))^{T} \\
    s.t.\,\,\omega ^*\left( \alpha \right) =\text{arg}\min_{\omega \in \Omega}\mathcal{L}_{train}\left( \omega ,\alpha \right)\\
    \alpha\in \varOmega_{\alpha}, \omega \in \varOmega_{\omega}
  \end{array}
  \label{equation_3_1}
\end{equation}
where $\varOmega_{\alpha} = \Pi^{n}_{i-1}[a_i,b_i]$ is the backbone architecture search space, and $a_i,b_i$ are the lower and upper bound. $\alpha = (\alpha_1,\dots,\alpha_n)^{T}\in \varOmega_{\alpha}$ is the candidate architectures. $\omega\in\varOmega_{\omega}$ is the lower-level variable and denotes its associated weights. The upper-level objective function $\mathbf{F}$ denotes the object detection loss $\mathcal{L}$ on the validation data $\mathcal{D}_{val}$, and the architecture complexity $\mathcal{C}_0$ and  $\mathcal{C}_1$. The lower level objective function $\mathcal{L}$ is the object detection loss on the training data $\mathcal{D}_{train}$.

Correspondingly, the process of the architecture search is performed according to Eq.~\ref{equation_3_1} and it can be simplified and shown in Fig.~\ref{fig_1}. The goal of the proposed method is to search a series of backbone architectures for object detection in mobile devices. Therefore, the structure of the backbone network will change continuously during the search (i.e., the way the nodes are linked in the backbone network are searched as shown in Fig.~\ref{fig_1}), while the number of input feature pyramid channels $C3,C4,C5$ is changed as the structure of the backbone network. The architecture of other parts of the object detection network remains constant. 

\begin{figure*}
  \centering  
  \includegraphics[width=14cm]{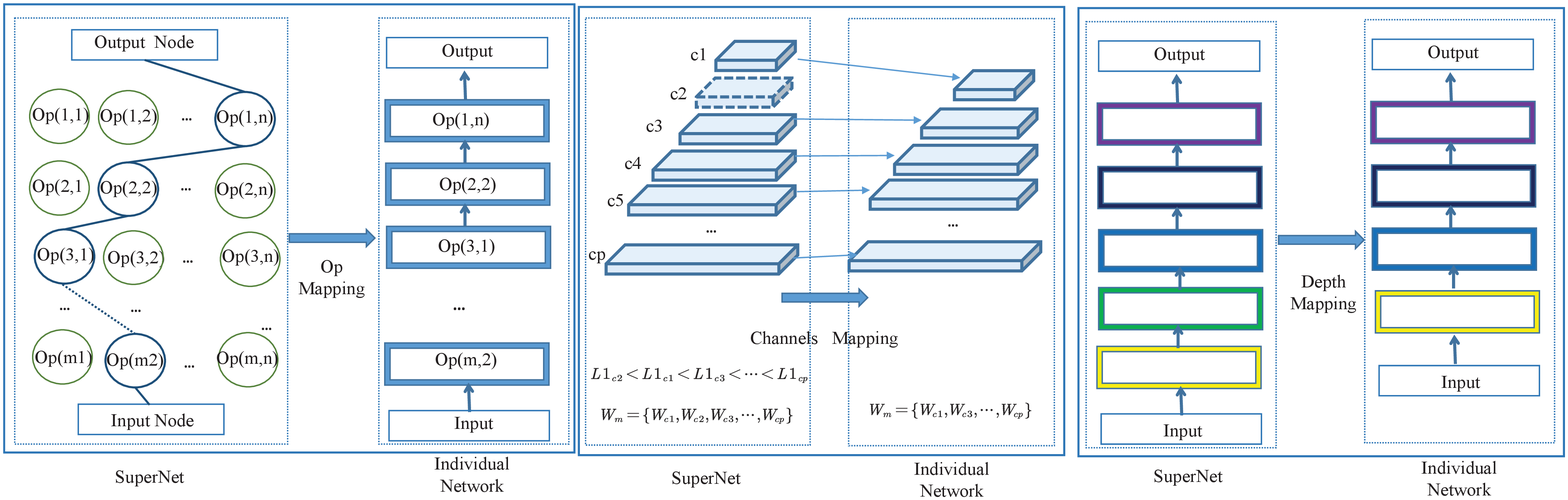}
  \caption{\footnotesize Weight mapping is consisted of three kinds of mapping: operation mapping, channels mapping, and depth mapping}
  \label{fig_2}
  \vspace{-2.0em}
\end{figure*}

Our proposed approach is a multi-objective evolutionary optimization of object detection backbone network architecture following SPOS \cite{guo2020single}. The method is performed in two main stages: the supernet pre-trained stage and the architecture search stage. In the supernet pre-trained stage, the supernet pre-trained on ImageNet is fine-tuned on COCO. Therefore, the fine-tuned supernet has the performance conditions to evaluate the backbone architecture on COCO. For the architecture search stage, the supernet is sampled as a single path one-shot approach with uniform sampling. The depth of the network is controlled by the number of identity operations. The number of output channels is searched by pruning via the L1 norm of the channel weights. We perform channel sorting and pick the important channels(with the larger L1 norm) to map the individual network weights. The network depth, network structure, and network out channels are encoded to be searched by the evolutionary optimization approach (shown in Algorithm \ref{20220509_1} ) based on non-dominated sorting for detection-specific NAS. During the search, each individual network architecture is sampled as a single path and a fixed channel width in the supernet via the weighting mapping technique. For the weighting mapping technique, the details can be seen in Section III-C.

In every iteration of MNSGA-NAS, a group of off-spring (new population) is created from the population by crossover and mutation applied to more promising already discovered structures, also known as parents. Each member of the population competes for survival and reproduction in each iteration and the initial population is randomly generated as shown in Algorithm \ref{20220509_1} line 1st. Subsequent to initialization, MNSGA-NAS conducted the search in two stages: 1) Fast non-dominated sorting for architecture complexity ($\mathcal{C}_1$,$\mathcal{C}_2$) that can be obtained quickly. In the first stage, the off-spring $Q_1$ is created through crossover and mutation with fast non-dominated sorting for architecture complexity ($\mathcal{C}_1$,$\mathcal{C}_2$) which can be obtained by a quick calculation. The off-spring $Q_1$ is also constrained by $g_1$and $g_2$. It is to increase the number of individuals in the constrained space, which can improve the efficiency of architecture search for mobile device scenarios. 2) for the second stage, the off-spring $Q$ is obtained through fast non-dominated sorting with  $ \left( \mathcal{L}_{val}\left( P \right) ,\mathcal{C}_1\left( P \right) \right),\mathcal{C}_2\left( P \right)$, which may consume a lot of computing resources because the process of pruning need to retrain the network to fast response accuracy. At the end of the evolution, a set of architectures reflecting an efficient trade-off between network performance and complexity is obtained through multiple learning processes of genetic operators. All the flowcharts and pseudocode outlining the MNSGA-NAS are shown in Fig.~\ref{fig_1} and Algorithm \ref{20220509_1}.

\subsection{Weight Mapping from Supernet}

In this section, we define weight mapping as a paradigm for mapping SuperNet weights to an individual network that represents the detection network formed during the architecture search process. We denote the individual network as $\mathbb{N}_{in}$ and the SuperNet as $\mathbb{N}_{super}$. The weight mapping is illustrated in the following three aspects: operation mapping, channel mapping, and depth mapping. The schematic diagram of weight mapping is shown in Fig.~\ref{fig_2}.

Supernet is with the max size channels, max size depth, and all kinds of operations. Each layer of the superNet represents the layer of the individual network. Thus, the operation in each layer of the individual network $\mathbb{N}_{in}$ can be mapped to the superNet $\mathbb{N}_{super}$. The weights of $i$th layer in supernet can be denoted as $\{o^{i,1}_{super},o^{i,2}_{super},\dots,o^{i,7}_{super}\}$, and the weights of $i$th layer in individual network can be denoted as $\{o^{i}_{in}\}$. The operation mapping process can be shown as the Op Mapping in Fig.~\ref{fig_2}. Operation mapping in $i$th layer is formulated as:

\begin{equation}
  \begin{array}{c}
    o^{i}_{in} = o^{i,j}_{super}, \ \   \forall  1 \leqq j \geqq 7
  \end{array}
  \label{equation_3_2}
\end{equation}

Channels mapping represents the process of channel pruning. For this reason, the Supernet is trained with full width and the network channels can be pruned flexibly. We calculate the channel importance based on the L1 norm of the channel weights, where a channel L1 norm means more importance. Therefore, the channel with a larger L1 norm is left during the channels pruning \cite{cai2019once,he2018amc}. As Eq.~\ref{equation_3_2}, we denote the channels of $o^{i,j}_{super}$ in supernet as $W^{(out,in,h,w)}_{super}\in \mathbb{R} ^{p\times q \times h \times w}$ and the channels of $o^{i}_{in}$ in individual network as $W^{(out,in,h,w)}_{in}\in \mathbb{R} ^{r\times s \times h \times w}$. $out,in$ denotes the output, input dimension of the operation and $h,w$ denotes the  spatial dimension. $p,q$ is the size channels of the supernet operation with the max size channels, and $p,q$ is the size channels of the operation of the individual network, where $   1 \leqq r \geqq p,  1 \leqq s \geqq q$. The channel mapping is illustrated as Fig.~\ref{fig_2}. The channels mapping process can be formulated as follows:

\begin{equation}
  \begin{array}{c}
    W^{i,j,:,:}_{in} = W^{i,j,:,:}_{super} \ \   \forall  1 \leqq r \geqq p, \forall  1 \leqq s \geqq q
  \end{array}
  \label{equation_3_3}
\end{equation}

Depth mapping is the process of adjusting the number of operations in the individual network. The operation of identity is utilized to control the depth of the individual network. As MobileNetV2 \cite{sandler2018mobilenetv2}, individual networks and supernet are split at some stage. We assume that one stage of the individual network is with $l$ layers and one stage of the supernet is with $m$ layers, where $l \leqq m $. The weight of one stage in an individual network can be denoted as $\{o_{in}^{1},o_{in}^{2},\dots,o_{in}^{l}\}$ and so as to the weight of one stage in supernet $\{o_{super}^{1},o_{super}^{2},\dots,o_{super}^{m}\}$. The depth mapping is illustrated in Fig.~\ref{fig_2} and the depth mapping process can be formulated as follows:

\begin{equation}
  \begin{array}{c}
    o^{i}_{in} = o^{i}_{super} \ \   \forall  1 \leqq l \geqq m,
  \end{array}
  \label{equation_3_4}
\end{equation}

\subsection{Mobile Search Spaces}

In the search strategy, we define the GhostNet \cite{han2020ghostnet} as the basic network and introduce more options of architecture elements. For every GhostNet-based module layer, we allow for kernel channels and more expansion size. For the adjustment of network depth, we control the depth of the network by adjusting the number of identity connections. For each layer, we offer 7 modules to choose from, and the GhostNet-based modules are as follows:

\begin{itemize}
  \item k3 GBe1: kernel size:$3\times3$, expansion size: 1. 
  \item k3 GBe2: kernel size:$3\times3$, expansion size: 2. 
  \item k3 GBe3: kernel size:$3\times3$, expansion size: 3. 
  \item k3 GBe4: kernel size:$3\times3$, expansion size: 4. 
  \item k3 GBe5: kernel size:$3\times3$, expansion size: 5. 
  \item k3 GBe6: kernel size:$3\times3$, expansion size: 6. 
  \item Identity
\end{itemize}

We set more layers in one stage of the backbone network. The structure of the search space is detailed in Table \ref{tab20220507}. Based on Table \ref{tab20220507} and the GhostNet-based modules, the proposed approach search for network depth, network structure and the size of the output channels. Because the obtained network needs to output features for detection, we set the third, fifth and sixth stages of the backbone to output the feature maps.

\begin{table}[htbp]
  \footnotesize 
  \centering
  \caption{ GhostNet-Based Search Space of Channels}
  \begin{threeparttable}
    \begin{tabular}{ccc}
      \toprule
      Stage & Channels & Feature Output \\
      \midrule
      1     & 16    & N \\
      2     & \{24,32\} & N \\
      3     & \{40,48\} & Y \\
      4     & \{56,64,72,80,88,96\} & N \\
      5     & \{104,112,120,128\} & Y \\
      6     & \{144,152,160,168,176,184,192\} & Y \\
      \bottomrule
      \end{tabular}%
    \begin{tablenotes}
      \footnotesize
      \item "Channels": the number of output channels. "Stage": the Stage of backbone network. "Feature Output": Y denotes the feature map corresponding to the output of in this stage and N denotes the opposite.
    \end{tablenotes}  
  \end{threeparttable}
  \label{tab20220507}%
  \vspace{-2.0em}
\end{table}%

\section{Experiments}

In the experiment, the supernet is pre-trained on ImageNet \cite{krizhevsky2017imagenet}. We randomly split the original training set into two parts: 50000 images are for validation (50 images for each class exactly) and the rest as the training set. ImageNet is mainly utlized to pre-trained supernet and backbone network,, and the image size is $224 \times 244$. The MS-COCO 2017 \cite{lin2014microsoft} is also utilized in the experiment. There are 118K images in the train set and 5K images in the Val set. This dataset is widely considered to be challenging, especially due to the large variation in object scale and a large number of objects per image. We report the average average accuracy (mAP) of object detection, and the image size is $320 \times 320$.

\subsection{Implementation details}

To further verify the effectiveness of the proposed method, we utilize the YoloX-Lite \cite{ge2021yolox} as the baseline. YoloX-Lite is the lighter version of YoloX \cite{ge2021yolox} which. The major difference between YoloX-Lite and YoloX is that YoloX-Lite is controlled the feature dimension of the head to be 64 so that it has similar FLOPs.  

\textbf{Supernet pre-training.} For the ImageNet classification dataset, we use the commonly used 1.28M training images for supernet pre-training. To train the one-shot supernet backbone on ImageNet, we set the batch size of 1024 on 8 GPUs for 300k iterations. The initial learning rate is set to be 0.5 and it is decreased linearly to 0. The momentum is 0.9 and weight decay is $4\times10^{-5}$.

\textbf{Supernet fine-truning.} In the step, the one-shot supernet is with detection head, feature fusion pyramid. YoloX is utilized as the base detector to be fused with the ImageNet pre-trained supernet. The training images are resized such that the shorter size is 320 Pixel. The network with supernet and detector is trained on 8 GPUs with a total of 16 images per minibatch for 140k iterations on COCO. The learned rate is 0.02. We use weight decay of $1\times10^{-5}$ and momentum of 0.9. 

\textbf{Search on supernet with non-dominated sorting.} The detection datasets are split into a training set for individual network retrained and a validation set for architecture search. For COCO, the validation set contains 5k images randomly selected from the COCO train set. Hyperparameters for training the individual network weights follow the short schedule for standalone training MMDetection. We use Adam optimizer with an initial learning rate of 0.2, weight decay of $1\times10^{-5}$, and momentum of 0.9. The individual networks are trained on 8 GPUs with a total of 16 images per minibatch for 7k iterations. The best candidates in the retrained individual networks.

\begin{table*}[htbp]
  \footnotesize 
  \centering
  \caption{\footnotesize Object Detection Results on MS-COCO. Test AP scores are based on COCO test-dev.}
  \begin{tabular}{ccccccccc}
    \toprule
    \multirow{2}[4]{*}{BackBone} & \multirow{2}[4]{*}{Input Res.} & \multicolumn{2}{c}{mAP} & \multirow{2}[4]{*}{Latency (ms)} & \multirow{2}[4]{*}{Device Name} & \multirow{2}[4]{*}{GFLOPs} & \multirow{2}[4]{*}{Madds} & \multirow{2}[4]{*}{Params} \\
\cmidrule{3-4}          &       & Valid & Test  &       &       &       &       &  \\
    \midrule
    MobileNetV2+MnasFPN \cite{chen2020mnasfpn} & 320   & 25.6  & 26.1  & 18.5  & EdgeTPU & -     & 0.92B & 2.5M \\
    MobileNetV2 + NAS-FPNLite  \cite{ghiasi2019fpn}
    & 320   & -     & 25.1  & 202   & CPU   & -     & 0.98B & 2.02M \\
    MobileNetV2 + SSDLite  \cite{ghiasi2019fpn}
    & 320   & -     & 22.1  & 200   & Pixel 1 CPU & 1.6   & -     & 4.3M \\
     MobileNetV2+FPNLite  \cite{RN1282} & 320   & -     & 24.3  & 264   & Pixel 1 CPU & 2.03  & -     & 2.2M \\
    MobileDets-IBN(EdgeTPU) \cite{RN1282} & 320   & 25.1  & 24.7  & 7.4   & EdgeTPU & -     & 0.97  & 4.17M \\
    YOLOv3-tiny \cite{redmon2018yolov3}  & 320   & 14    & -     & 71    & Galaxy S20 & 3.3   & -     & 8.85M \\
    YOLObile (GPU only) \cite{cai2021yolobile} & 320   & 31.6  & -     & 58.8  & Galaxy S20 & 3.95  & -     & 4.59M \\
    YOLO-ReT-M0.75(INT8) \cite{ganesh2022yolo} & 320   & 18.4  & -     & 30.12 & Jetson Nano & -     & -     & 5.2M \\
    YOLO-ReT-M1.4(INT8) \cite{ganesh2022yolo} & 320   & 19.1  & -     & 43.45 & Jetson Nano & -     & -     & 12.3M \\
    YOLO-ReT-EB 3(INT8) \cite{ganesh2022yolo} & 320   & 19.7  & -     & 91.24 & Jetson Nano & -     & -     & 28.3M \\
    \midrule
    GhostNet+YoloX Lite & 320   & 24.6  & 24.1  & 17.23 & Jetson TX2 NX & 0.74  & 0.70B & 3.56M \\
    SPOS CLS.+YoloX Lite & 320   & 26.6  & 26.8  & 17.35    & Jetson TX2 NX & 0.997 & 0.96B & 3.16M \\
    \midrule
    MNSGA-NAS v1+YoloX & 320   & 26.1  & 26.3  & 15.46 & Jetson TX2 NX & 0.75  & 0.72B & 1.96M \\
    MNSGA-NAS v2+YoloX & 320   & 26.4  & 26.9  & 16.53    & Jetson TX2 NX & 0.83  & 0.81B & 2.64M \\
    MNSGA-NAS v3+YoloX & 320   & 27.6  & 27.7  & 17.18 & Jetson TX2 NX & 0.971 & 0.95B & 2.36M \\
    \bottomrule
    \end{tabular}%
  \label{tab:2}%
  \vspace{-2.0em}
\end{table*}%

\textbf{Training from scrath.} As the backbone architecture searching is finished, we train the network as that batch size is 16 images per GPU and the SGD optimizer with momentum 0.9, learning rate 0.8, and weight decay 0.0001. The learning rate scheduler configuration is set to decay as the number of epoch steps, the detection networks are trained for 300 Epochs. As a while, Mixup and Mosaic implementation are utilized to train out models and close them for the last 15 epochs. For strong data augmentation, ImageNet pre-training is no more beneficial and even leads to overfitting. So all models are trained from scratch on detection datasets.

\textbf{Device Setup.} For fairness comparison, we compare the runtime inference latency of various detection models on Jetson TX2 NX and the input image resolution is set to 320*320. All the detection models are optimized by TensorRT 8.2 as FP16 to speed up execution. We target real-time execution with batch size =1, so we process 20 images to warm up the inferencing and 100 images to test the latency. We take the average execution time with 100 images for inference latency calculation. We also utilize mmdeploy \cite{mmdeploy} to transform the mmdetection-based \cite{mmdetection} model for TensorRT-based inference.

\subsection{Main results}

We compare the architectures obtained via MNSGA-NAS against state-of-the-art mobile detection models on COCO. For a fair comparison, the result is reported with mean average precision (mAP) and floating-point operations per second (FLOPS) and compared with other mobile detection models. Results are represented in Table~\ref{tab:2}. As the similar-level FLOPs, we find that searching directly on the detection task (MNSGA-NAS v1+YoloX) allows us to improve mAP by 1.5 mAP over GhostNet+YoloX Lite on the validation set and 2.8 mAP on COCO test-dev.

For other state-of-the-art mobile detection models, we compare with the NAS-crafted MobileDet \cite{RN1282}, NAS-FPNLite \cite{ghiasi2019fpn} and MnasFPN \cite{chen2020mnasfpn} and achieve competitive results. MNSGA-NAS v3 outperforms MobileDets-IBN \cite{RN1282} by 3.0 mAP on COCO test-dev as comparable FLOPs. Likewise, MNSGA-NAS v2 outperforms MobileNetV2+SSDLite  \cite{ghiasi2019fpn} by 4.8 mAP with smaller than 2X FLOPs. Similarly, MNSGA-NAS v1 also outperforms MobileNetV2+MnasFPN by 0.5 mAP on COCO valid set as comparable FLOPs. The results confirm the effectiveness of detection-specific NAS.

To exclude the influence of the search space and YOLOX framework on the performance, the results also report the performance of the optimal individual (SPOS CLS.+YoloX Lite) based on the ImageNet classification task search with the same search space. The backbone is obtained by following the SPOS \cite{guo2020single}, the population size is set to 20 and the number of iterations is 50. Likewise, SPOS CLS.+YoloX Lite is not with ImageNet pre-training. As can be seen that MNSGA-NAS v3 outperforms SPOS CLS.+YoloX Lite by 1.0 mAP on COCO valid set as comparable FLOPs. The results show that MNSGA-NAS again shows effectiveness in detecting dedicated NAS compared to architecture search for classification tasks.

These results support the fact that network architecture search on mobile, task-oriented can make the network more inclined towards high performance. With our non-dominated sort-based algorithm, we allow us to achieve a better trade-off between floating-point operations per second and the accuracy of the model.

\subsection{Architecture Visualizations}

\begin{table}[htbp]
  \vspace{-1.0em}
  \footnotesize
  \centering
  \caption{\footnotesize The architectures of MNSGA-NAS v3}
    \begin{tabular}{ccccc}
    \toprule
     No.  & Opertion & Input Chan. & Outnput Chan. & Stride \\
    \midrule
    0     & Conv3 & 3     & 16    & 2 \\
    \midrule
    1     & K3GBe3 & 16    & 16    & 2 \\
    2     & K3GBe2 & 16    & 32    & 1 \\
    3     & K3GBe4 & 32    & 32    & 1 \\
    4     & K3GBe6 & 32    & 32    & 1 \\
    \midrule
    5     & K3GBe2 & 32    & 40    & 2 \\
    6     & K3GBe3 & 40    & 40    & 1 \\
    7     & K3GBe5 & 40    & 40    & 1 \\
    8     & K3GBe3 & 40    & 40    & 1 \\
    \midrule
    9     & K3GBe2 & 40    & 56    & 2 \\
    10    & K3GBe1 & 56    & 56    & 1 \\
    11    & K3GBe4 & 56    & 56    & 1 \\
    12    & K3GBe3 & 56    & 56    & 1 \\
    13    & K3GBe4 & 56    & 56    & 1 \\
    \midrule
    14    & K3GBe6 & 56    & 128   & 1 \\
    15    & K3GBe6 & 128   & 128   & 1 \\
    16    & K3GBe1 & 128   & 128   & 1 \\
    17    & K3GBe1 & 128   & 128   & 1 \\
    18    & K3GBe1 & 128   & 128   & 1 \\
    19    & K3GBe1 & 128   & 128   & 1 \\
    20    & K3GBe2 & 128   & 128   & 1 \\
    21    & K3GBe3 & 128   & 128   & 1 \\
    22    & K3GBe5 & 128   & 128   & 1 \\
    \midrule
    23    & K3GBe3 & 128   & 152   & 2 \\
    24    & K3GBe3 & 152   & 152   & 1 \\
    25    & K3GBe6 & 152   & 152   & 1 \\
    26    & K3GBe4 & 152   & 152   & 1 \\
    27    & K3GBe4 & 152   & 152   & 1 \\
    28    & K3GBe5 & 152   & 152   & 1 \\
    29    & K3GBe3 & 152   & 152   & 1 \\
    30    & K3GBe6 & 152   & 152   & 1 \\
    31    & K3GBe6 & 152   & 152   & 1 \\
    32    & K3GBe4 & 152   & 152   & 1 \\
    33    & K3GBe3 & 152   & 152   & 1 \\
    \bottomrule
    \end{tabular}%
  \label{tab:3}%
  \vspace{-2.0em}
\end{table}%

\begin{figure*}
  \vspace{-0.5cm} 
\setlength{\belowcaptionskip}{-1.5cm} 
  \centering
  \subfigure[The individuals distribution of backbone FLOPS and loss for NSGA-II and MNSGA-NAS]{
      \begin{minipage}[b]{0.35\textwidth}
      \includegraphics[width=0.9\textwidth]{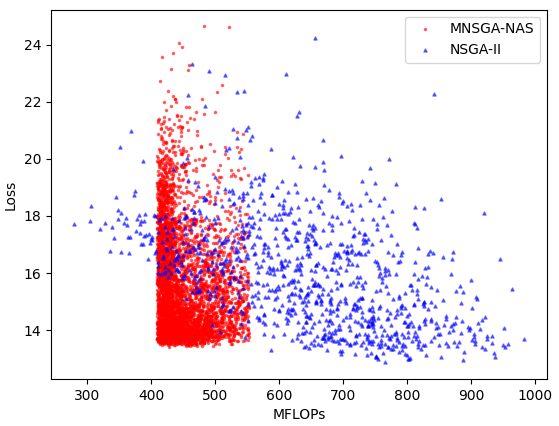}
      \end{minipage}
  }
  \subfigure[The individuals distribution of backbone FLOPS, parameters and loss for NSGA-II and MNSGA-NAS]{
      \begin{minipage}[b]{0.35\textwidth}
      \includegraphics[width=0.9\textwidth]{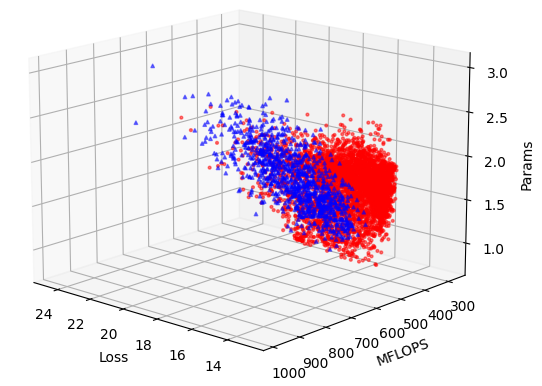}
      \end{minipage}
  }
  \caption{\footnotesize  The individuals distribution for NSGA-II and MNSGA-NAS} \label{fig:2}
  \vspace{-2.0em}
\end{figure*}

Table~\ref{tab:3} illustrates our searched object detection backbone architecture, MNSGA-NAS v3. 'No.' indicates the layers of the backbone. Meanwhile, the backbone needs to provide the output feature maps of multi-scale so it is divided into 6 stages. Layer 0 denotes the first stage, the second stage contains layers 1-4, and 5-8,9-13,14-22, and 23-33 denote the third, fourth, fifth, and sixth stages, respectively. Where the output feature maps of the third, fifth and sixth stages are treated as the feature outputs of the backbone network. An interesting observation is that the number of output channels does not increase as double when the size of the feature map is changed. The results reveal that the adjustment of output channels for the backbone is necessary to improve the model performance. The relationship between the number of backbone network output channels and the data may require further analysis.

\subsection{Ablation Study}

In this section, we reveal the effectiveness of the proposed method by performing a series of ablation experiments. We first visualize the distribution of individuals during the search process to show the difference between the evolutionary process of MNSGA-NAS and that of NSGA-II. In addition, we also compare the advantages of the proposed approach in terms of accuracy, the number of parameters, and inference time as comparable FLOPs, which further demonstrates the comparative advantages of our algorithm.

MNSGA-NAS is improved based on NSGA-II with the characteristics of NAS. Analyzing the distribution of individuals of both algorithms can better reveal the fundamental principle of MNSGA-NAS. Fig.~\ref{fig:2}-(a) shows the distribution of individuals in the search process for MNSGA-NAS and NSGA-II. The blue dots and red dots indicate the individuals of MNSGA-NAS and NSGA-II in the search process, respectively. In addition, the horizontal axis indicates the FLOPs of the backbone and the vertical axis indicates the YoloX Lite loss of the corresponding individual as the backbone. Lower YoloX loss indicates higher accuracy of the backbone, so the individual with lower loss is more preferred to the desired architecture at the same FLIOPs. From Fig.~\ref{fig:2}-(a), it can be seen that for the same FLOPs, red points with lower losses than blue points exists, and the red points are the ones we need. The results reveal that MNSGA-NAS optimizes NSGA-II based on non-dominated sorting for the NAS scenario. NAGA-II is computationally solved for the complete Pareto front surface, so there will be a more uniform portion of individuals on the Pareto surface. However, NAGA-II acquired several unwanted individuals, causing a decrease in search efficiency. On the other hand, MNSGA-NAS performs non-dominated sorting twice within the constraint space, which allows for a finite number of iterations, and searches for individuals with a lower size of FLOPs and lower loss. Fig.~\ref{fig:2}-(b) adds the distribution of the parameter sizes. Fig.~\ref{fig:2}-(b) shows the Pareto front surface of the MNSGA-NAS search in three dimensions, which is more in line with the needs of low loss, low FLOPs, and low parameters of NAS. It further explains the advantages of MNSGA-NAS for NAS problems.

\begin{table}[htbp]
  \scriptsize
  \centering
  \caption{The experimental results comparing NSGA-II and MNSGA-NAS in similar floating-point operation states }
    \begin{tabular}{cccccc}
    \toprule
    \multirow{2}[4]{*}{Backbone} & \multicolumn{2}{c}{GFLOPs} & \multirow{2}[4]{*}{Latency(ms)} & \multicolumn{2}{c}{mAP} \\
\cmidrule{2-3}\cmidrule{5-6}          & Backbone & All   &       & \multicolumn{1}{l}{Valid} & \multicolumn{1}{l}{test-dev} \\
    \midrule
    NSGA-II Based & 0.529   & 0.993   & 18.35 & 27.1  & 27.3 \\
    MNSGA-NAS v3 & \textbf{0.525}   & \textbf{0.971}   & \textbf{17.06} & \textbf{27.8}  & \textbf{27.9} \\
    \bottomrule
    \end{tabular}%
  \label{tab3}%
\end{table}%

In addition, we chose the individual (NSGA-II Based) from NSGA-II Pareto surface individuals with FLOPs similar to those of MNSGA-NAS v3. The NSGA-II Based inference time is measured on the same Jetson TX2 NX device, and the performance comparisons are shown in Table~\ref{tab3}. The results in Table~\ref{tab3} show that MNSGA-NAS v3 outperforms NSGA-II Based by 0.6 mAP on COCO test-dev and 0.7 mAP on COCO Valid Set as comparable FLOPs. MNSGA-NAS v3 also improve 1.29ms than NSGA-II Based on latency. These results support the fact that the individuals obtained by MNSGA-NAS search have an advantage over NSGA-II in terms of inference time and prediction accuracy. The results in Table~\ref{tab3} further support the conclusions in Fig.~\ref{fig:2}. 

\section{Conclusion}

In this work, we address the problem of suboptimal results caused by a linear combination of objectives in multi-objective neural architecture search, propose a non-dominated ranking-based evolutionary algorithm for NAS scenarios, and use it for backbone network architecture search for object detection tasks. This method designs a GhostNet-based search space for mobile devices. The proposed approach automatically designs the backbone network from the network depth, network structure, and network output channels via a technique of weight mapping, making it possible to use NAS for mobile devices detection tasks a lot more efficiently. It follows SPOS and fuses a pre-trained supernet as a backbone into YoloX, fine-tuned based on the COCO dataset. Finally, the backbone network is obtained through the evolutionary search by the proposed approach. Experiments show that the architectures produced by the proposed approach achieve excellent detection results on the mobile hardware platform, which are significantly better than the state-of-the-art to a large margin. The results also report that the proposed approach achieves higher accuracy at comparable FLOPs compared to the classification task-oriented SPOS, again showing the effectiveness of detecting dedicated NAS.


%
\IEEEpeerreviewmaketitle




\bibliographystyle{IEEEtran}

\bibliography{cas-refs}

\end{document}